**Bridging battery design and health assessment through virtual sensing and physics-informed learning**

Wendi Guo[1]*, Søren Byg Vilsen[2], Daniel Ioan Stroe[3], Yaqi Li[3], Yicun Huang[4], Ashima Verma[5], Daniel Brandell[1]*

*[1] *Department of Chemistry, Uppsala University, Sweden*

[2] *Department of Materials and Production, Aalborg University, Denmark*

[3] *Department of Energy, Aalborg University, Denmark*

[4] *Department of Information Technology, Uppsala University, Sweden*

[5] *Alstom Rail Sweden AB, Sweden*

*e-mail: wendi.guo@kemi.uu.se, daniel.brandell@kemi.uu.se

**Abstract**

Supercharging of lithium-ion batteries (LiBs) requires robust health monitoring to ensure durability, safety and user confidence, particularly for emerging vehicle-to-grid applications with bidirectional energy flows. Yet battery management remains largely disconnected from the material and structural origins of aging, limiting both interpretable health assessment and informed battery design. Here we propose a physics-informed learning framework with virtual sensing that infers hard-to-measure design parameters, including solid-state diffusion coefficient, electrode thickness, ion concentration, and particle size, directly from standard battery management system (BMS) measurements. Across diverse fast-charging strategies and driving profiles, embedding a digital-twin-derived particle-cracking mechanism as a soft constraint reduces trajectory and lifetime prediction errors by 6–8 times relative to state-of-the-art machine learning baselines using only 2% early-life observations. We further show that accurate degradation extrapolation does not require fully resolved governing equations; validated partial mechanisms, jointly refined with limited data, provide sufficient guidance. Virtual sensing transforms standard charging signals into latent design variables without additional sensors, bridging observable battery behavior and underlying aging processes while reducing capacity loss error by up to 39%, end-of-life (EOL) error by 17%, and prediction variability by up to 54%, enabling real-time exploration of new battery configurations. More broadly, the proposed framework establishes a practical feedback loop between deployment and development, demonstrating how real-world operation can continuously inform upstream design decisions across complex multiphysics systems.

## Introduction

Fast charging is becoming essential for modern electric-vehicles (EVs) and emerging vehicle-to-grid (V2G) services[1], improving user experience while motivating charging strategies and battery designs capable of sustaining long-term operation. However, high charging rates accelerate degradation, creating an urgent need for accurate lifetime prediction[2], early diagnosis[3], and interpretable health assessment. Without insights into the underlying aging mechanisms, battery management systems (BMSs) lose interpretability for reliable decision-making[4], while battery development remains disconnected from application requirements before large-scale deployment. Yet owing to coupled multiscale physical and chemical processes, establishing a direct link between chemistry and material-design parameters to real-world performance remains a major challenge[5].

Numerical simulation[6] serves as a primary route for connecting intrinsic design factors (such as chemistry, material properties, and geometry) with practical usage conditions (such as electric loading, SOC range, and temperature). In particular, multiphysics modeling enables quantitative aging analysis and design-performance evaluation, forming the foundation of battery digital twins[7]. One method is electrochemical-mechanical coupling, where lithiation-induced stress drives particle cracking[8] and continuous SEI growth on newly exposed graphite surfaces. These processes consume cyclable lithium while imprinting aging signatures onto voltage[9], impedance[10], power[11], and capacity[11,12]. Another method is electrochemical–thermal coupling, where nonuniform temperature fields regulate local reaction kinetics[13], while spatially heterogeneous current densities drive heat generation and thermal instability[14]. Thermal measurements thereby reveal how charging current[15] interact with the chemical processes for the specific electrode architecture to influence safety and lifetime[16]. A more comprehensive method couples electrochemical, mechanical and thermal effects through weak[17] or strong[12] interactions to jointly capture charging behavior, aging and thermal response across scales. These high-fidelity models establish a physics-based connection between battery design and field performance, supporting both operational improvement and upstream design optimization[18]. However, their high computational cost and reliance on solving nonlinear partial differential equations (PDEs) prevent real-time prediction, adaptive control and inverse inference of material properties, while generally remaining incompatible with the high-dimensional streaming data generated by modern BMSs.

These limitations have driven an increasing interest in machine learning (ML). ML naturally provides tools for extracting informative features from high-dimensional sensing data[19] and building predictive models for health and safety management[20]. With continuous monitoring of current[3,], voltage[3,9], temperature[21], and impedance[22,23] together with electrochemical diagnostic indicators[24] such as incremental capacity (IC)[25] and differential voltage (DV)[26], large volumes of multi-fidelity data are now available for battery diagnosis and prognostics. However, pure ML models lack mechanistic observability and therefore cannot infer the underlying material and structural properties that govern aging behavior, such as ionic diffusivity[27], electrode geometry[28] and particle size[29]. These latent design factors[30] remain difficult to identify from standard BMS measurements alone. There is a clear gap in connecting intrinsic battery design with observable behavior. Bridging this gap requires combining the feature extraction capability of ML with the physical interpretability of multiphysics models, without requiring additional sensors in manufacturing and deployment.

In this context, physics-informed learning[31] has emerged as a promising strategy to achieve this integration by embedding domain knowledge into ML through governing physical rules[32]. Two major routes have developed[32,33]: data-assisted physical models and physics-guided data-driven models. Most efforts focus on physically meaningful observables derived from experiments[34] or simulations[35], such as half-cell open-circuit-voltage (OCV) curves[36,37], lithium plating potentials[35], and shifts in IC peak positions or areas[24,38]. These approaches have improved prediction accuracy for battery management[39]. In contrast, virtual sensing, defined here as converting standard BMS signals into latent physical and design-state estimations without additional sensors, remains largely unexplored for linking observable battery behavior with underlying aging mechanisms and cell design. Physics-informed neural networks (PINNs) represent one of the prominent frameworks within this paradigm [40]. Existing studies have demonstrated improved health and lifetime prediction by incorporating degradation-related equations into neural models[41,42]. However, current PINN implementations rely primarily on empirical constraints, whereas explicitly validated mechanisms are rarely embedded into the learning process[9,41]. More importantly, PINNs are often treated as predictive tools rather than a means to understand how constitutive laws shape neural learning and inject physical intuition into optimization. Understanding this process is important for AI-driven scientific discovery[43], where generating scientific insight may be as important—or more important—as improving prediction accuracy itself. Further exploring the potential of PINNs for interpretable prediction, adaptive control, and deployment-informed design represents a promising direction for integrating battery development with field operation in next-generation energy-storage systems.

To address this gap and bridge battery design with real-world application requirements, we propose a physics-informed learning framework with a virtual sensing strategy for both rapid prototype screening and onboard health assessment. This framework infers latent design parameters, including solid-state diffusion and electrode architecture, directly from existing EV sensing signals. Particle-cracking dynamics derived from a digital twin (DT) model[44] are incorporated through soft physical constraints, enabling accurate degradation extrapolation under limited data availability. The framework is developed using two fast-charging datasets based on graphite/$LiNi_{0.5}Mn_{0.3}Co_{0.2}O_2$ (NMC532) LiB cells, covering charging rates from 1C to 2C, four multistep fast-charging protocols, dynamic discharge profiles spanning 70–100% depth of discharge (DOD), and four ambient temperatures (0, 15, 25 and 35°C) representative of EV operating conditions in Nordic climates[45]. The proposed framework achieves trajectory and lifetime prediction errors 6–8 times lower than state-of-the-art ML baselines across previously unseen operating profiles. Notably, stable extrapolation is achieved using only 2% observations, corresponding to fewer than 19 equivalent full cycles (EFCs) for the longest-lifetime cells, below both the widely adopted 100-cycle benchmark[46] and recent 50-cycle early-life results[47]. Our findings further reveal that accurate degradation extrapolation does not require complete mechanistic knowledge. Instead, validated partial physical mechanisms, jointly refined with data, are sufficient to recover aging behavior. Incorporating virtual sensing of the negative-electrode diffusion coefficient, thickness, and particle size reduces capacity loss error by up to 39%, EOL error by 17%, and prediction variability by up to 54%. Once trained, the framework replaces repeated multiphysics simulations with millisecond-level physics-informed computation (<0.13s versus 210,996 s for a single COMSOL aging simulation), enabling rapid exploration of battery design and operating spaces. The inferred design parameters help open the "black box" linking observable signals to hidden aging mechanisms, providing interpretable and actionable guidance for battery design.

## Results

### Virtual sensing of latent design states

Our aim is to move beyond purely statistical ML battery prognostics by linking real-world operating profiles with latent design states. This connection enables both mechanistic interpretation of degradation behavior and design-oriented feedback for battery development. To achieve this, we identify health indicators that capture diverse aging behaviors (Table S1). Building on previous findings[4], four key design parameters ($D_{neg}$, $L_{neg}$, $R_{neg}$, and $C_{s,pos}$) are virtually sensed through a validated DT model (Figure 1). In this work, virtual sensing refers to estimating physically interpretable P2D design parameters from partial charging segments using standard BMS signals and DT-generated training data. These parameters correspond to material and geometric properties of the cell rather than data-driven latent features, with their physical interpretation and measurement methods summarized in Supplementary Table S3. Synthetic datasets are generated by perturbing these parameters within a 1% relative standard deviation to emulate high-precision manufacturing variability, revealing latent electrochemical signatures associated with electrode geometry and ionic diffusion coefficient that are difficult to access experimentally. As illustrated in Figure 1, voltage, capacity, capacity variation and time are extracted from short charging segments (highlighted by the red rectangle). By learning the electrochemical-response manifold induced by varying design states, the framework enables real-time inverse inference of transport and structural parameters directly from random charging segments. In this way, virtual sensing shifts battery diagnostics from monitoring external responses to tracking the hidden upstream design information governing degradation evolution and failure susceptibility[14].

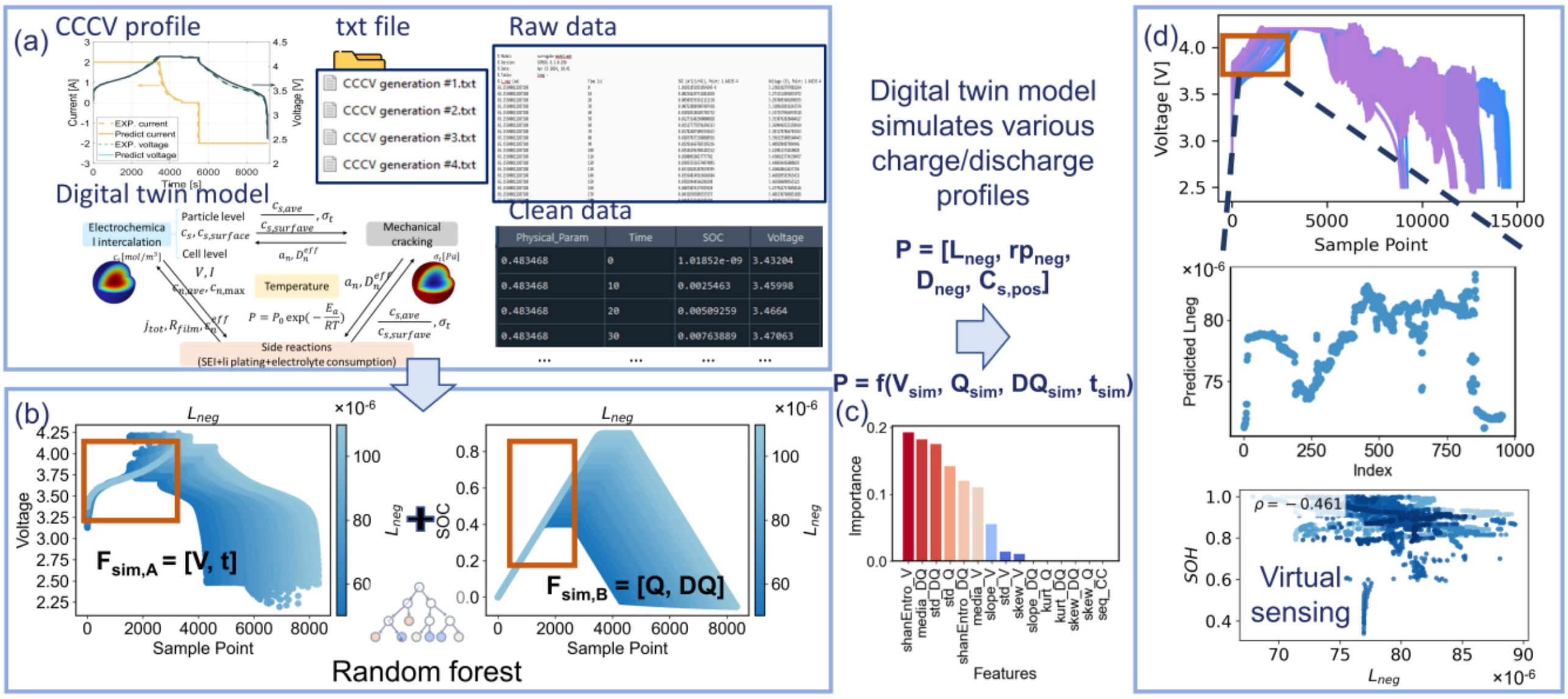

Figure 1. Virtual sensing from partial charging segment. (a) The DT model is used to vary four key design parameters and generate diverse charge/discharge profiles. (b) Mapping between partial charging responses and latent design parameters, illustrated using the inferred negative electrode thickness ($L_{neg}$). Features are extracted from voltage, capacity, capacity variation and charging-time descriptors. (c) The middle panel shows the corresponding feature-importance ranking for $L_{neg}$ inference, where $f(\cdot)$ maps charging-segment features to latent design parameters that encode aging information. (d) Real-time inference of design parameters from dynamic operating profiles. The bottom panel illustrates the relationship between $L_{neg}$ and SOH.

This study utilizes fast-charging datasets based on 18650 cylindrical graphite/NMC532 cells with a nominal capacity of 2 Ah. The datasets comprise a publicly available fast-charging dataset[4] together with an additional dataset, including multistep charging protocols designed to emulate dynamic EV duty cycles (Supplementary Note 1). Experimental conditions are summarized in Table S2 and Figure

S1, with further details provided in the respective reference papers[44,48,49]. Except for two cells cycled at 1.3C@15 °C, each cell was operated under a unique combination of charging/discharging protocol and thermal condition, resulting in highly heterogeneous degradation trajectories. To the best of our knowledge, this is the first EV-oriented battery-aging dataset exploring multistep fast charging, dynamic discharge operation and temperature variability within a unified cycling framework representative of driving conditions in Nordic climates. During cycling, current (I), voltage (V), capacity (Q), and cycle index are continuously recorded at fixed sampling intervals across both constant-current and multistep charging stages.

To account for protocol-dependent voltage windows, equivalent full cycles (EFCs) are defined by normalizing cumulative discharged capacity to the nominal cell capacity, providing a more realistic representation of EV usage than raw cycle counts alone[50]. Sixteen health indicators are extracted from randomly sampled charging segments to characterize temporal and internal electrochemical behavior through non-invasive IC/DV responses (Table S1). Their mechanistic interpretation is detailed in Ref.[4]. To further capture latent design states, virtual-sensing variables are inferred from partial charging segments. Figure 2 summarizes the evolution of both measured health indicators and inferred design states under heterogeneous operating conditions.

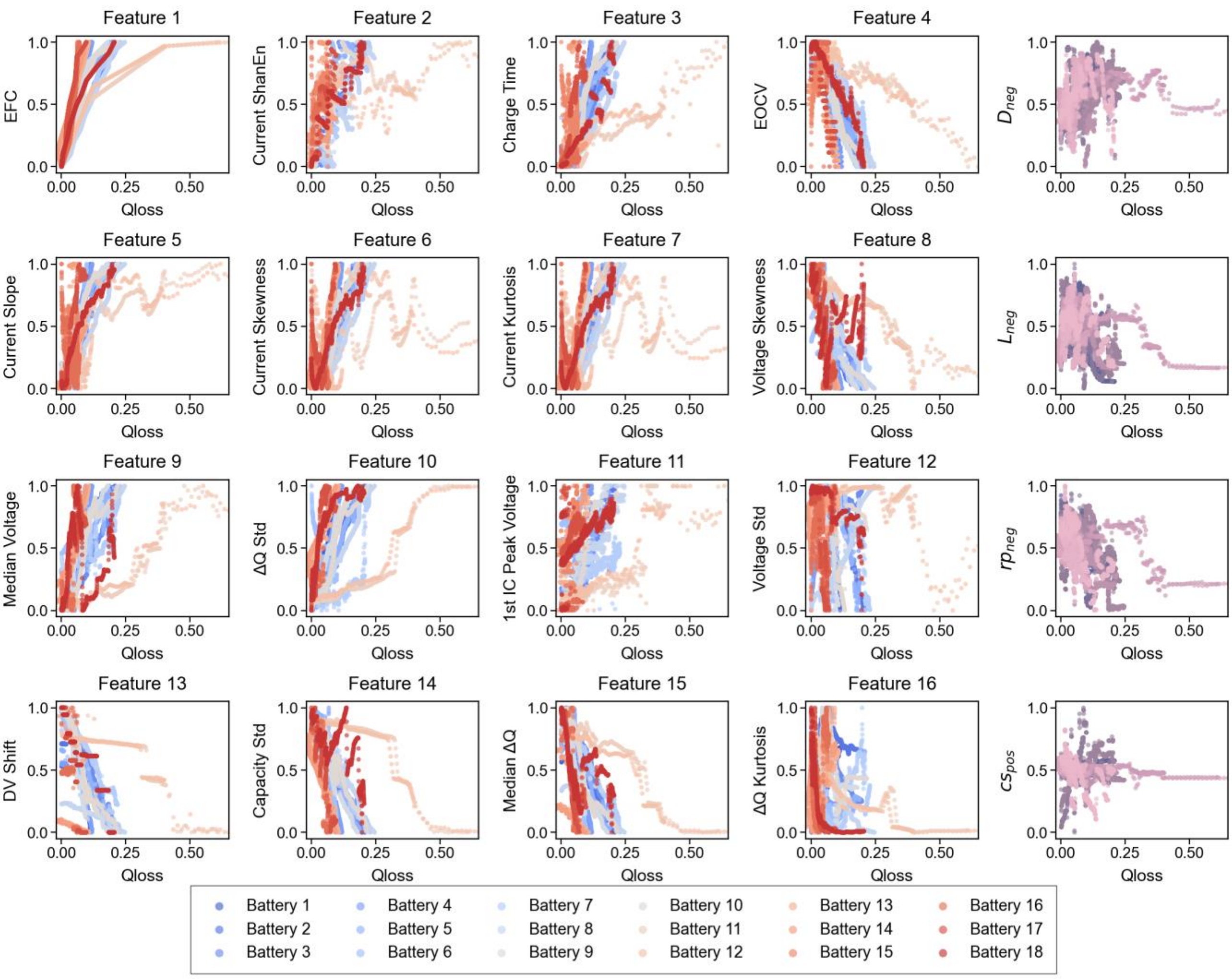

Figure 2. Evolution of health indicators and inferred design variables from partial charging segments. Sixteen engineered features together with three inferred design parameters are mapped against capacity loss ($Q_{loss}$). In each subplot, the features have been normalized by its magnitude under the restricted charging window ($V_l$=3.65V and $V_u$=4.10V). Lastly, the color gradients denote the different cells from two fast charging datasets.

**Physics-informed learning framework**

Traditional ML methods learn correlations between observable signatures and degradation trajectories rather than the physics governing battery aging, fundamentally limiting generalization under unseen conditions and limited early-life observations. This challenge arises from two factors: (i) heterogeneous operating conditions activate different coupled aging mechanisms, making transferable representations inherently difficult to learn; and (ii) early-stage measurements contain only weak and partially developed failure precursors, providing insufficient information for reliable long-term extrapolation without physical constraints. We propose a physics-informed learning framework in which particle-cracking dynamics are incorporated into the loss function via soft penalty constraints. To the best of our knowledge, this is the first framework to integrate explicit electrochemical equations derived from validated DT knowledge into neural-network training. The inclusion of particle cracking is motivated because Ni-rich NMC cathodes undergo severe chemo-mechanical instability during high-voltage cycling. Increasing Ni content enhances energy density but also induces large anisotropic lattice contraction at high delithiation states[51], accelerating crack propagation and interfacial degradation. Importantly, particle cracking involves both gradual and accelerated aging regimes[44], providing a compact mechanistic representation of heterogeneous degradation behavior. To bridge incomplete constitutive physics and evolving cell-level aging characteristics, we further introduce a two-stage adaptive physics-informed training strategy.

As illustrated in Figure 3, the proposed framework is explicitly designed for strict extrapolation rather than interpolation within randomly shuffled trajectories. Unlike existing PINN studies[41], which often relax temporal causality during train–test split, the present framework preserves chronological aging evolution throughout optimization and extrapolation processes. Source cells are sequentially split into training and validation subsets to establish a global physics-constrained solution manifold, whereas target cells are restricted to the first 2–20% of the full trajectory for adaptation and early stopping (see Supplementary Note 2 for details of data splitting strategy). All subsequent states remain completely hidden during training and are used exclusively for future-state prediction, making the task substantially more challenging and physically meaningful than standard interpolation-based benchmarks.

In Phase I, source cells with sufficiently matured aging states (typically ≥50% observable capacity loss) initialize the global constitutive structure. During this stage, the physical parameters $\theta$ remain fixed while the PDE residual is enforced over dense collocation points spanning the full EFC domain. Initial-condition anchoring and monotonicity regularization further constrain physically admissible trajectory formation. Consequently, the network learns globally consistent degradation behavior governed by the embedded particle-cracking law rather than isolated feature–label correlations. Phase II performs adaptive transfer using only the first 2–20% of the target degradation trajectory. Starting from the pretrained physics-constrained manifold, the framework exposes only limited early-stage measurements to the network, while all future states remain fully unobserved. The physical parameters are unfrozen and co-optimized with the neural-network weights using decoupled learning rates, enabling continuous recalibration to unseen thermal, charging and driving conditions. A differentiable positivity reparameterization guarantees physically admissible parameter updates, whereas a physics-integrity sentinel iteratively rejects solutions exhibiting trajectory reversal, numerical instability or implausible future-state inference (see Methods for details).

The framework is motivated by two practical scenarios: early-stage battery prototype screening, where only limited initial cycling data are available to assess long-term design viability, and onboard battery-management systems (BMSs), where future degradation must be inferred online from sparse historical observations. In both cases, the objective is not interpolation within densely sampled degradation trajectories, but physically consistent prediction of future aging evolution and EOL directly from short charging segments under EV dynamic profiles.

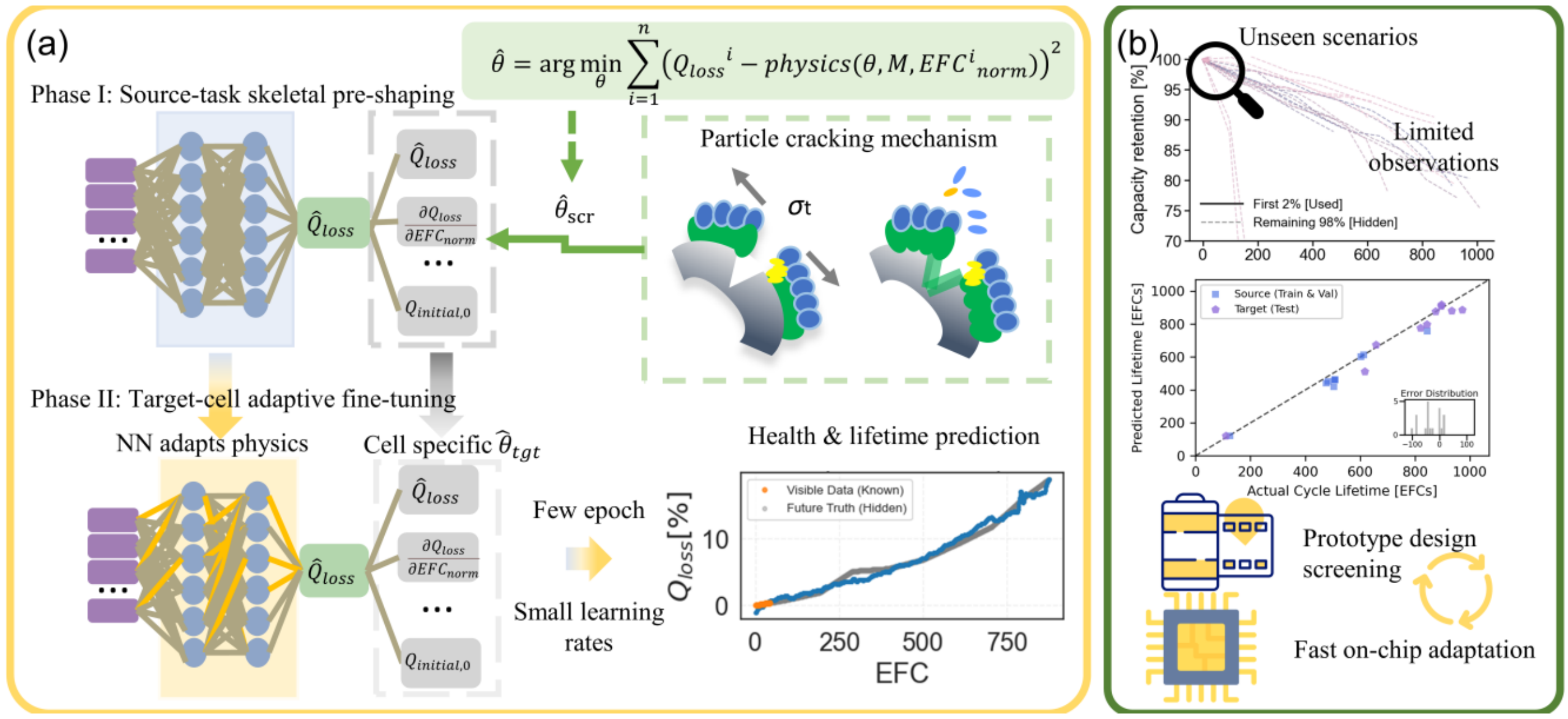

Figure 3. A physics-informed learning pipeline for battery health and lifetime prediction using only 2% observations. (a) In-silico PINN training with the particle-cracking law embedded into the optimization objective as a soft constitutive constraint. The neural-network weights and physical parameters $\boldsymbol{\theta}$ are jointly optimized to satisfy both data and physical constraints and reproduce degradation trajectories. (b) Deployment with only the first 2% of observations for full-trajectory and end-of-life extrapolation. The framework enables rapid prototype screening and interpretable onboard health assessment for real-world battery-management systems.

**Co-evolution of neural and constitutive dynamics**

Using a representative unseen multistep fast-charging cell under driving and partial V2G operation, we monitor the evolution of the physical loss terms during PINN optimization to reveal the underlying learning behavior (Figure 4a). The governing-equation residual decreases from ~$1.4\times10^{-4}$ to ~$1.0\times10^{-4}$ over 80 epochs, while the derivative-consistency term remains near ~$3\times10^{-3}$, showing that optimization gradually transitions from local fitting toward physics-governed degradation dynamics (Figure 4(a-1)). The weighted physics penalty simultaneously decreases from $4.7\times10^{-2}$ to $2.9\times10^{-2}$, demonstrating progressively stronger agreement between the neural solution and the embedded particle-cracking constraint (Figure 4(a-2)). Throughout optimization, the monotonicity penalty remains zero, indicating that capacity loss is intrinsically preserved without explicit trajectory correction. Meanwhile, the zero-anchor constraint stabilizes the initial degradation regime within the ~$10^{-10}$–$10^{-5}$ range, suppressing nonphysical capacity recovery during extrapolation (Figure 4(a-3)). Figure 4b further visualizes parameters adaption during transfer refinement. The degradation-sensitive parameters $A_u$ and $Z_u$ increase monotonically from approximately 0 to 1.7%, whereas $B_u$ decreases to −1.3%, revealing progressive rebalancing of the embedded aging kinetics under limited early-cycle supervision. Physically, $A_u$ is associated with material-dependent crack-growth scaling[44], whereas $Z_u$ governs diffusion-induced tangential stress and crack-depth propagation, indicating adaptive modulation of stress-coupled fracture behavior. By contrast, $\Phi$, associated with SEI density, molar mass and layer thickness, remains nearly invariant, suggesting that the pretrained PINN captures the dominant Solid Electrode Interphase (SEI)-growth regime without requiring substantial parameter recalibration (see Methods for details).

We further open the "black box" of PINN optimization by visualizing the physics-constrained loss landscape, revealing how embedded constitutive laws inject physical intuition into neural learning (see Supplementary Note 3). Underweight perturbations, the proposed PINN forms a sharply confined loss basin with a well-defined minimum below $\log_{10}(\text{MSE})\approx-3.6$, whereas the FNN, CNN, and LSTM baselines exhibit substantially flatter landscapes centered around higher errors of $\approx-2.7$, $\approx-1.9$, and $\approx-2.5$, respectively (Figure 4c). The pronounced directional curvature and narrow valley geometry of the PINN indicate stronger optimization confinement and improved parameter identifiability during future-state inference. By contrast, the shallow and weakly structured surfaces of the ML baselines suggest higher solution degeneracy and greater sensitivity to parameter perturbations. These observations provide a mechanistic explanation for the superior stability of the proposed PINN when inferring unseen states. Optimization landscapes for the remaining held-out test cells are provided in Figure S3-S11. Finally, we examined whether the learned neural responses preserve the embedded aging physics throughout optimization for the representative unseen cell (Figure 4d). The predicted $Q_{loss}$ closely follows the particle-cracking trajectory across the normalized EFC domain (Figure 4(d-1)), with the neural and physics-constrained solutions remaining nearly indistinguishable up to $\text{EFC}/M_{phys}\approx0.45$, where $M_{phys}$=1200 normalizes EFC into the [0,1] interval for all cells (see Supplementary Note 4). The learned gradients $\partial y/\partial EFC_{norm}$ simultaneously reproduce the local crack-growth kinetics, capturing the progressive SEI coverage of freshly exposed electrode surfaces and the associated incremental capacity loss[10] (Figure 4(d-2)). Agreement in both state and derivative space demonstrates that the proposed PINN internalizes the underlying constitutive behavior governing their evolution.

These results show that effective physics-informed learning does not require a fully resolved electrochemical model or perfectly known governing equations. Instead, incorporating a subset of validated physical relationships and refining them together with data provides sufficient constraints for accurate degradation prediction, even under unseen operating conditions and heterogeneous aging behaviors.

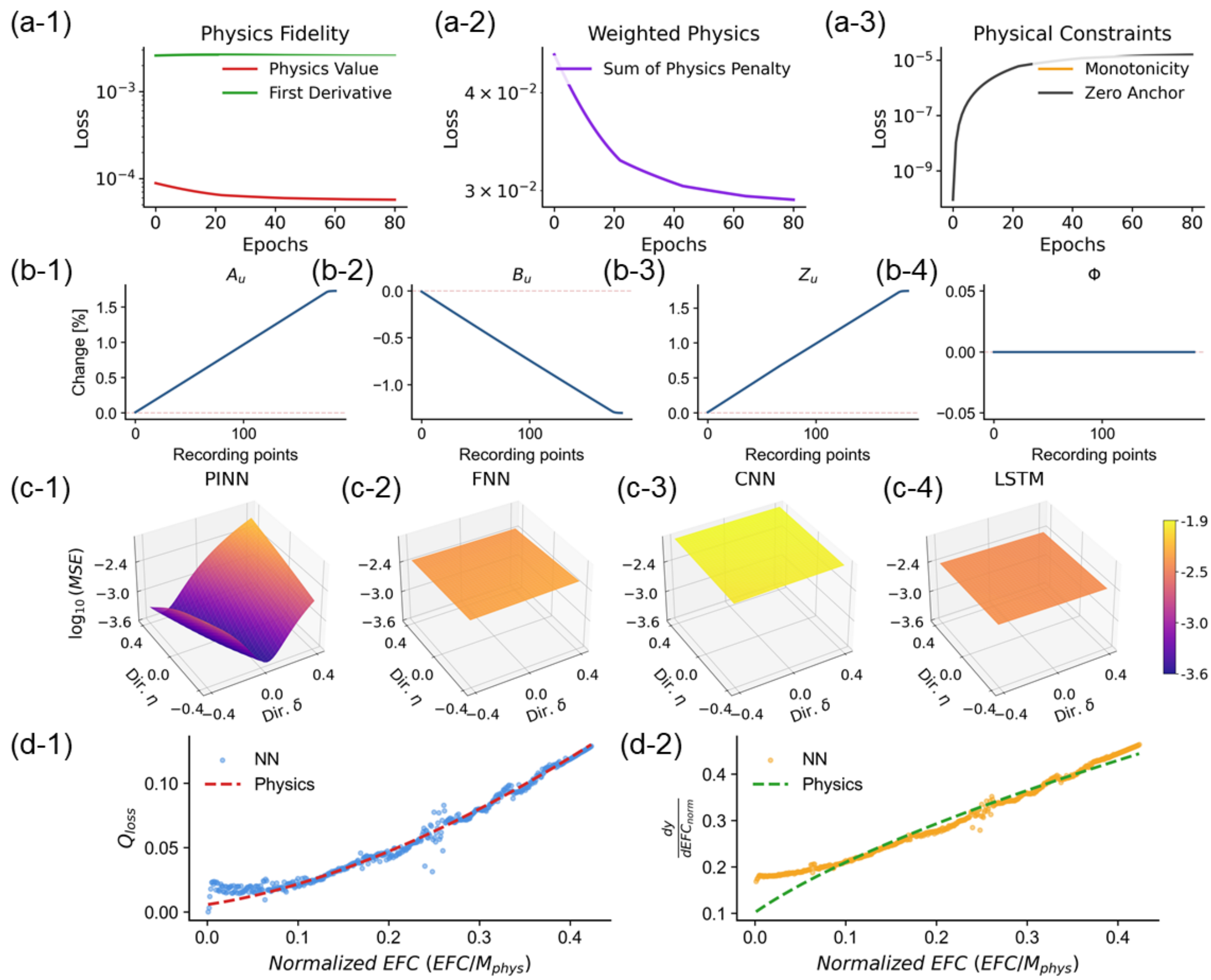

Figure 4. Neural–constitutive co-evolution during PINN optimization. (a) Evolution of physics-constrained losses for a representative unseen fast-charging cell under dynamic driving and partial V2G operation, including governing-equation fidelity, derivative consistency, weighted physics penalties, monotonicity regularization, and zero-anchor constraints. (b) Adaptive refinement of the latent physical coefficients during transfer optimization under sparse observations. (c) Physics-constrained loss landscapes under weight perturbations. The proposed PINN forms a sharply confined loss basin with strong directional curvature, whereas the FNN (c-2), CNN (c-3), and LSTM (c-4) baselines exhibit substantially flatter optimization geometries. (d) Mechanistic consistency between neural responses and embedded aging physics.

**Sparse data and stable extrapolation**

Physics-informed learning gains robustness not by memorizing aging trajectories, but by constraining neural optimization to physically admissible solutions. This reduced solution space enables stable inference under highly limited observations. To evaluate this low-data inference capability, we conduct systematic experiments across varying training and transfer conditions.

We progressively increase the source-training ratio from 50% to 80% while keeping all target cells fully unseen. During adaptation, only the first 20% of each target trajectory is exposed for parameter refinement, with validation loss serving as the early stopping criterion (see Supplementary Note 7). Figure 5a demonstrates the strong data efficiency of the proposed PINN. With only 50% source-history availability, the PINN achieves a trajectory MAE of 6%, whereas the FNN, CNN and LSTM baselines remain near 10.5%, 9.2% and 11.0%, respectively. This corresponds to relative error reductions of 35–50%. Increasing the source-history ratio to 80% further reduces the MAE to below 5%, while the ML baselines remain above 8.7%, 7.8% and 9.2%, respectively. In addition to lower median errors, the PINN also consistently exhibits narrower error distributions across all training conditions. Figure S12 further shows that the PINN preserves stable capacity loss extrapolation, suggesting strong potential for data-constrained onboard adaptation and real-world parameter calibration. The advantage becomes more pronounced for EOL inference (Figure 5(a-2)). Across all training conditions, the proposed PINN maintains stable relative lifetime errors of 20–25%, with only weak dependence on the amount of available data. By contrast, the FNN, CNN, and LSTM baselines exhibit large errors of 130–145%, 110–140%, and 125–145%, respectively, together with markedly wider prediction distributions. Increasing the source-training ratio from 50% to 80% further reduces the PINN EOL error from 24% to 20%, whereas all purely ML baselines remain above 110–160% across every training scale. Overall, the proposed PINN reduces trajectory and EOL errors by 40–50% and 5–7 times, respectively, despite using only 44% of the cells and 50% of the observable trajectory.

Nest, we set the source-training ratio to 80% and expose only minimal target-domain measurements to probe how far the PINN can project future degradation. The most challenging setting uses only the first 2% of the degradation trajectory, corresponding to just 19 EFCs for the longest-lifetime cell, where measurable aging signatures are still barely distinguishable (Figure S1). Despite this limitation, the proposed PINN achieves a trajectory MAE of 5%, whereas the FNN, CNN and LSTM baselines remain near 9.5%, 8.8% and 10.2%, respectively (Figure 5(b-1)). Increasing the visible trajectory ratio from 2% to 20% lowers the baselines errors to 8.2%, 7.0% and 8.3%, respectively. By contrast, the PINN remains confined within the 4–6% range across all adaptation conditions, showing that embedded physical laws sharply restrict the space of admissible degradation pathways, such that even sparse early-life observations are sufficient to determine future evolution. The difference becomes particularly evident under sparse adaptation conditions (Figure 5(b-2)). In all settings, our PINN achieves distinct EOL error reductions as compared to the other models. For instance, the ML baselines collapse when applied to unseen aging patterns with only the first 2% target observations (Figure S15). In contrast, the proposed PINN remains within 15–22%, corresponding to 6–8 times lower prediction errors. This improvement arises from the embedded constitutive constraints, which enable physics-guided transfer even under incomplete mechanistic knowledge and sparse observations.

Finally, we compare the predicted aging trajectories under the most challenging 2% adaptation across 10 random initialization seeds (see Supplementary Note 4). The proposed PINN accurately reconstructs the degradation manifold up to ~40% capacity loss, with errors tightly concentrated within ±5% around zero and minimal seed-to-seed variability. Although the CNN achieves the best baseline performance (Figure 5c-2), its trajectories progressively saturate beyond ∼10% capacity loss and fail to capture the accelerated aging regime. The corresponding error distribution becomes strongly skewed and expands beyond ±15%, together with substantially larger stochastic variability across random initializations. This discrepancy directly explains the large lifetime-prediction gap between the proposed PINN and ML baselines. Additional results under different observation ratios are provided in Figure S13. Furthermore, we find that combining 80% source-cell training with only 2% target-cell observations provides an effective balance between inference efficiency and robust prediction.

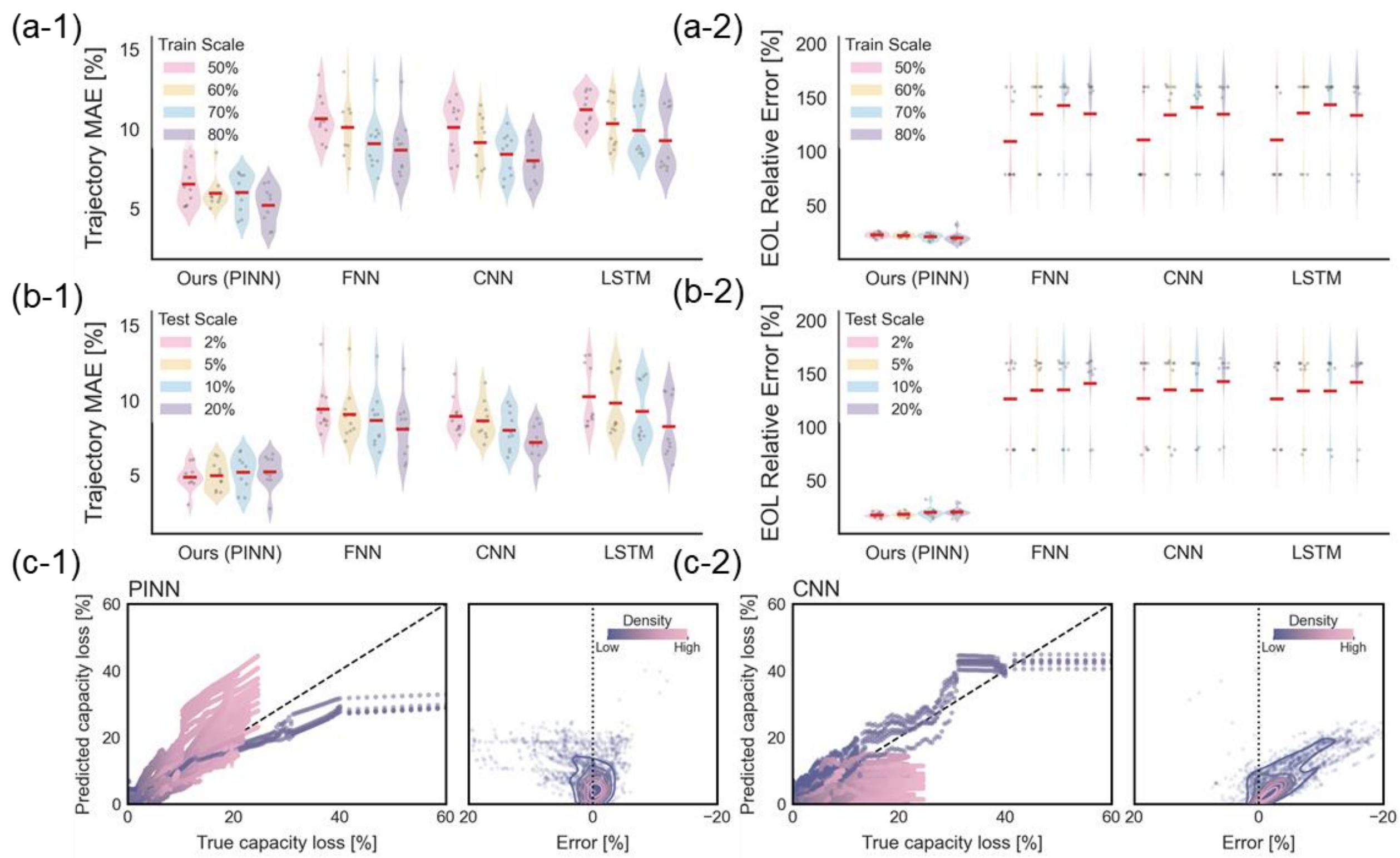

Figure 5. Physics-guided degradation and EOL extrapolation from limited early-life observations. (a) Effect of training-data availability on trajectory reconstruction (a-1) and EOL prediction (a-2) across unseen cells, with models trained using 50–80% of the available source-cell data. (b) Effect of sparse target observations during adaption. Only 2–20% of the initial target trajectory was provided during refinement. (c) Future degradation reconstruction under the extreme 2% observation condition across 10 random initialization seeds for the proposed PINN (c-1) and the best-performing CNN baseline (c-2).

**Impact of virtual sensing on PINN robustness**

To improve degradation prediction under realistic operating conditions, four key design parameters[4] are introduced into the PINN through virtual sensing. This strategy enables the model to access hidden design information from standard EV sensing signals, providing an additional physical layer beyond observable charging behavior (Figure 6).

Figure 6(a-1) shows the average SHAP attribution across all unseen cells over 10 random initialization seeds. Besides the dominant contribution of EFC (0.08–0.10), the inferred diffusion coefficient consistently exhibits SHAP values of 0.01–0.02, comparable to major voltage, differential-capacity and charging-time descriptors. Under multistep fast-charging operation (Figure 6(a-2)), the diffusion coefficient becomes even more influential, with lower diffusion-related SHAP values consistently associated with reduced capacity loss, highlighting diffusion-limited transport as a key driver of aging[52]. Incorporating the virtually sensed diffusion coefficient substantially reshapes the extrapolation landscape, forming a smoother and more symmetric optimization basin (Figure 6(b) and Supplementary Note 5 for details). Without this transport property, the loss surface remains sharply folded and highly asymmetric (Figure 6(b-1)). The diffusion-constrained PINN develops a well-confined minimum below $\log_{10}(\mathrm{MSE}) \approx -6.0$ together with substantially reduced directional sensitivity under weight perturbations (Figure 6(b-2)). The resulting loss landscape geometry strongly suppresses unstable local minima during extrapolation.

Figure 6c reveals heterogeneity in the inferred diffusion trajectories, with normalized diffusion coefficients spanning 0.1–1.1 over the lifetime range. The contribution of the diffusion coefficient to degradation prediction varies substantially across cells, with some exhibiting rapid early-to-midlife increases followed by pronounced late-stage decline, whereas others remain comparatively stable under different charging strategies and operating conditions. Incorporating the adaptive diffusion parameter further improves lifetime extrapolation under the extreme 2% condition (Figure 6d). The average cycle-life prediction $R^2$ increases from 0.63 ± 0.17 to 0.75 ± 0.11, while the MAE decreases from 71.27±16.07 EFCs to 60.55±9.73 EFCs. The predicted lifetimes align more closely with the parity line, particularly beyond 600 EFCs. Correspondingly, the error distribution becomes substantially narrower and more centered around zero, with markedly fewer outliers exceeding ±200 EFCs (Figure 6(d-3)). Mechanistically, such chemical diffusion heterogeneity[53] likely arises from spatially nonuniform Li ion flux under dynamic EV operation, where local variations in temperature[13], pressure[12], and electrochemical driving force promote irregular interfacial degradation[54] and accelerated capacity loss.

To further assess the influence of the remaining high-impact design parameters, we sequentially incorporate the virtual-sensing parameters following the order: $L_{neg}$, $R_{neg}$, $C_{pos}$. Incorporating virtual sensing consistently improves both capacity loss and lifetime prediction accuracy, with the largest performance gain observed after introducing $R_{neg}$ (Table 1). SHAP analysis further shows that larger negative-electrode particle radius is consistently associated with positive contributions to capacity loss (Figure S23), indicating that longer diffusion pathways intensify transport limitation and stress accumulation during fast charging. The inferred $L_{neg}$ similarly exhibits a positive correlation with degradation, suggesting amplified transport heterogeneity within thicker electrodes (Figure S20). By contrast, $C_{pos}$ produces weaker but still measurable SHAP contributions (Figure S26), implying additional coupling between local lithium concentration gradients and electrolyte accessibility[52].

Compared with the baseline PINN, incorporating latent design parameters reduces the capacity loss MAE by up to 39% ($R_{neg}$) and the EOL error by 17%, while increasing training time by less than 0.5 s. Prediction variability is also consistently decreased, with the standard deviation reduced by 28% ($C_{pos}$) to 54% ($L_{neg}$), demonstrating that virtual sensing improves predictive robustness with marginal computational overhead. Although DT data generation (≈4h for 1200 samples) and PINN training (≈ 4s) are one-time costs, even the slowest virtual-sensing configuration completes degradation and lifetime prediction in <0.13s versus ≈210,996s for a single COMSOL aging simulation (Figure S32), representing an acceleration of nearly six orders of magnitude. This efficiency enables rapid screening of new cell designs and material parameters across diverse operating profiles without repeated multiphysics simulations.

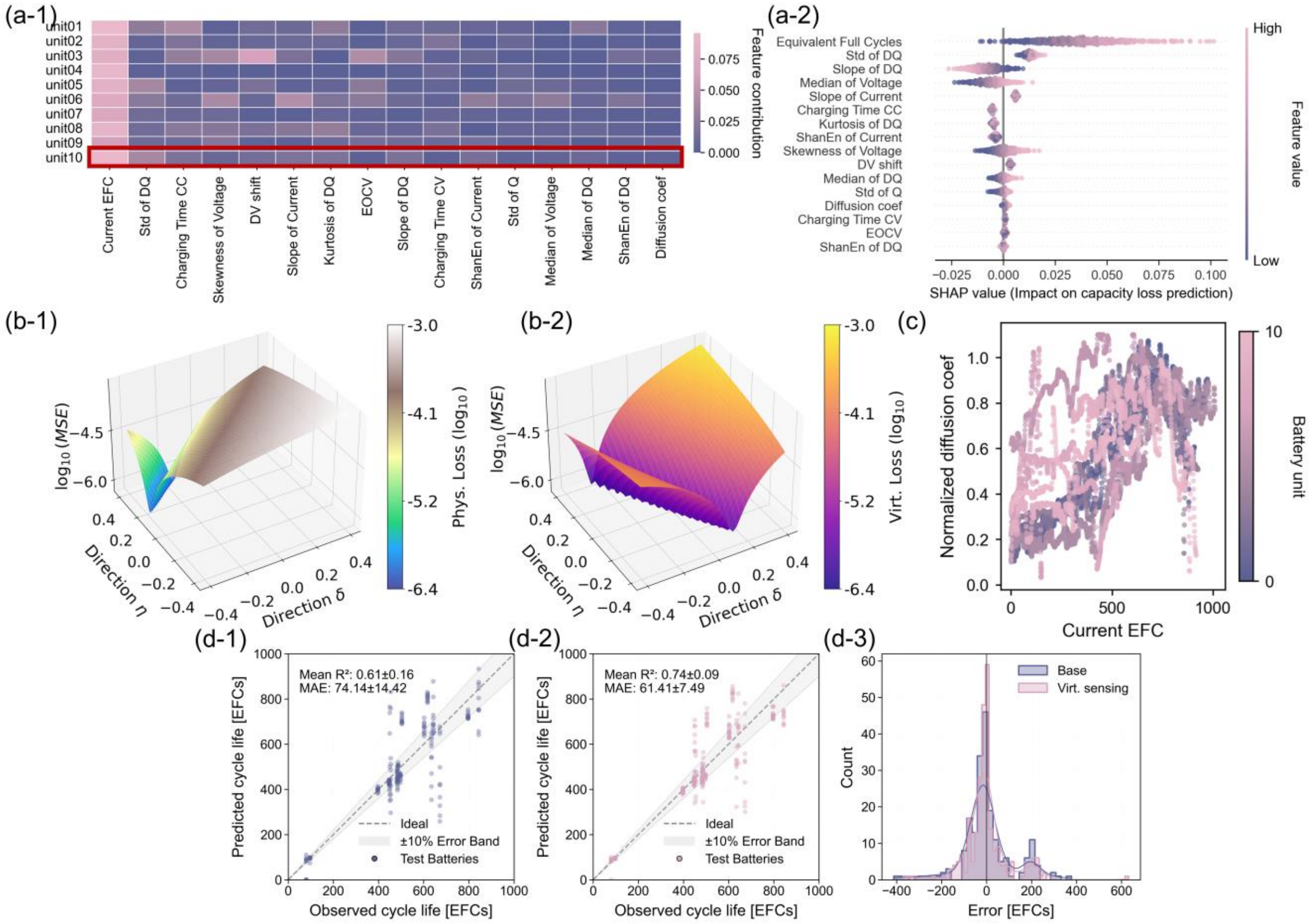

Figure 6. Impact of Li ion diffusion coefficient on degradation and EOL extrapolation. (a) SHAP analysis of the inferred diffusion coefficient across unseen cells. (a-1) Cross-cell SHAP attribution. (a-2) SHAP summary distribution under multistep fast-charging operation. (b) Extrapolation loss landscapes under weight perturbations. (b-1) PINN without diffusion embedding. (b-2) Diffusion-constrained PINN. (c) Evolution of the normalized diffusion coefficient, revealing heterogeneous

transport trajectories during aging. (d) EOL extrapolation performance with and without diffusion embedding. (d-1) Diffusion-constrained PINN. (d-2) Baseline PINN. (d-3) Distribution of EOL prediction errors.

**Table 1** Test mean absolute errors (MAE) and computational cost for the PINN and baselines.

| Models | #Features | Design parameters | Capacity Loss [%] | | | End-of-life [EFCs] | | | Train Time [s] | Inference Time [s] |
|---|---|---|---|---|---|---|---|---|---|---|
| | | | MAE | STD | $R^2$ | MAE | STD | $R^2$ | | |
| PINN only | 16[m] | No | 1.87 | 0.14 | 0.68 | 74.14 | 15.20 | 0.60 | 3.80 | 0.11 |
| PINN+$D_{neg}$ | 15[m]+1[d] | Yes | 1.26 | 0.20 | 0.85 | 61.41 | 7.89 | 0.75 | 4.21 | 0.13 |
| PINN+$L_{neg}$ | 15[m]+1[d] | Yes | 1.22 | 0.20 | 0.87 | 59.39 | 9.88 | 0.78 | 3.89 | 0.11 |
| PINN+$R_{neg}$ | 15[m]+1[d] | Yes | 1.24 | 0.18 | 0.89 | 59.15 | 12.38 | 0.77 | 4.32 | 0.12 |
| PINN+$C_{pos}$ | 15[m]+1[d] | Yes | 1.25 | 0.21 | 0.86 | 61.28 | 12.26 | 0.76 | 4.14 | 0.12 |

[m] manually engineered features

[d] virtual sensing parameters

**Aging inference under V2G charging constraints**

V2G operation restricts battery monitoring to short charging segments[55], creating a challenging scenario for aging prediction. To assess the proposed framework under these constraints, we compare PINN with and without virtual sensing against purely data-driven baselines across narrow charging windows and diverse fast-charging strategies. The most demanding case uses only 2% early-life observations from unseen cells, corresponding to 2–19 EFC checkpoints across different EV duty cycles.

Figure 6a compares capacity loss prediction under the narrow charging window (3.80−4.10 V). Virtual sensing produces tighter alignment with the parity line and suppresses large deviations during late-stage aging. The corresponding error distribution contracts from approximately ±13% to ±12% with markedly higher density near zero, indicating improved stability under constrained sensing conditions. To further assess the influence of underlying design states on health prediction, SHAP analysis is performed on the narrowest charging window (3.80−3.85 V) for one multistep fast-charging profile (Figure 6b). This window corresponds to the typical ~50% SOC regime used during bidirectional V2G operation. Incorporating virtual sensing introduces clear chemical attribution otherwise absent in the baseline PINN. Specifically, the inferred diffusion coefficient ($D_{neg}$) exhibits

SHAP contributions of ~0.01–0.02, indicating that the $Li^+$ transport dynamics remain strongly relevant to capacity loss even under highly restricted charging observations (Figure S30).

Figure 6c compares three purely ML baselines together with the proposed PINN with and without virtual sensing under the extreme 2% condition. All results reflect the average performance across all unseen cells within the narrowest charging window (3.80−3.85 V). The proposed PINN with virtual sensing achieves the lowest MAE/RMSE (2.20%/2.84%), outperforming the baseline PINN (3.07%/3.95%) and reducing the error nearly twofold relative to the best-performing ML model. This improvement arises because the key physical properties, particularly the diffusion coefficient within active particles, inherently regulate charging kinetics and overpotential accumulation[52]. Figures 6d compares trajectory prediction robustness across different partial charging windows. Without virtual sensing, the PINN remains highly sensitive to voltage-window selection, with several restricted ranges exceeding 5.5% MAE. Incorporating virtual sensing produces substantially more uniform accuracy, keeping most charging windows below 3% MAE, including the highly constrained [3.80 V, 3.85 V] range. Consequently, the fraction of voltage windows achieving MAE below 3% increases from 81.0% to 85.7%. Figure 6e further evaluates EOL extrapolation across the same voltage-window space. Without virtual sensing, multiple restricted windows exceed 180–200 EFC error. In contrast, virtual sensing constrains most lifetime errors within 80–140 EFCs despite the narrow sensing range, increasing the fraction of windows below 100 EFCs from 38.1% to 45.2%. These results highlight the importance of virtual sensing for practical V2G deployment, where battery health and plug-in eligibility must often be inferred from short and incomplete charging segments rather than full cycling trajectories [55].

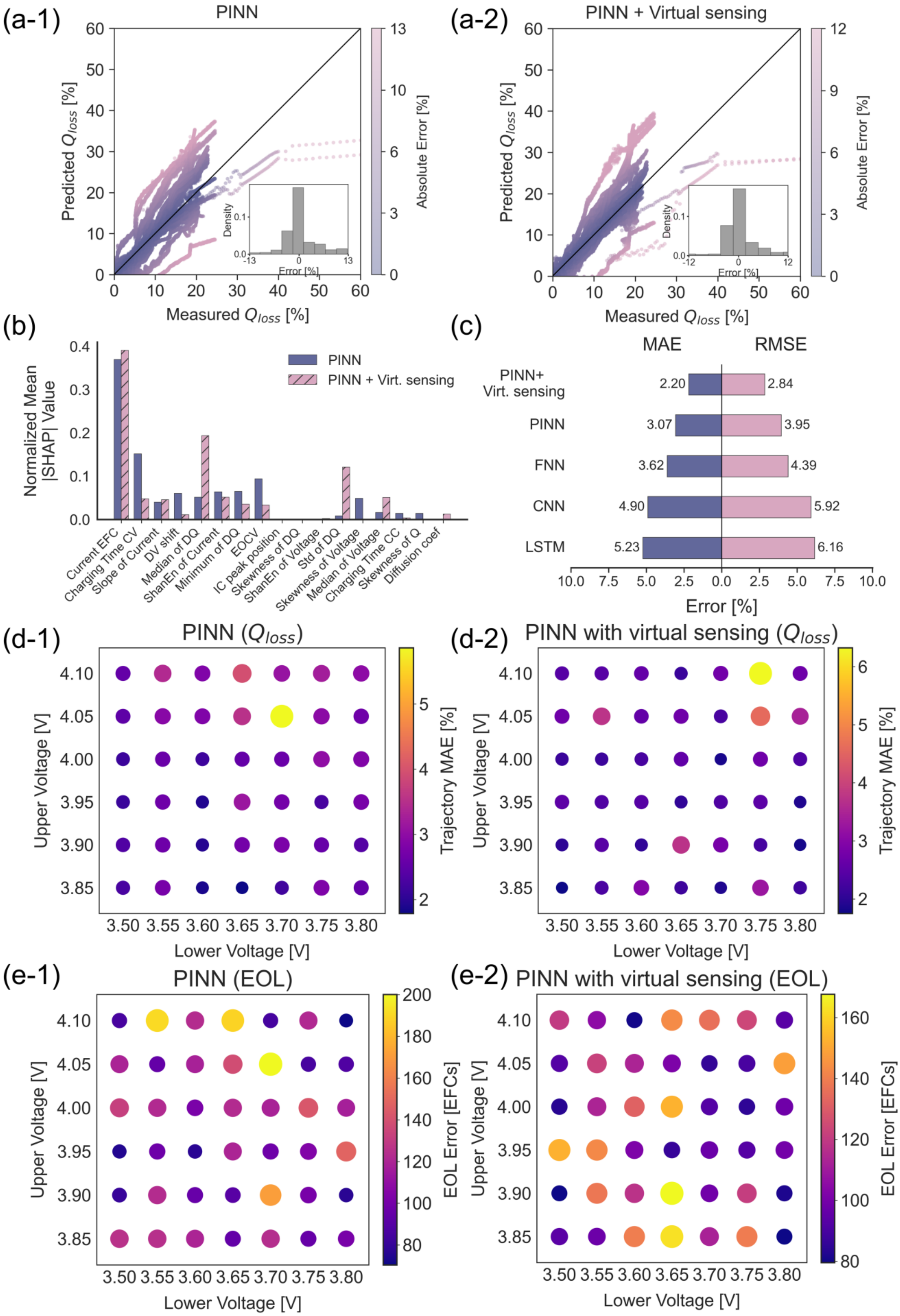

(a-1)
PINN
Predicted $Q_{loss}$ [%]
Measured $Q_{loss}$ [%]
Absolute Error [%]
Density
Error [%]
(a-2)
PINN + Virtual sensing
(b)
Normalized Mean |SHAP| Value
PINN
PINN + Virt. sensing
Current EFC
Charging Time CV
Slope of Current
DV shift
Median of DQ
ShanEn of Current
Minimum of DQ
EOCV
IC peak position
Skewness of DQ
ShanEn of Voltage
Std of DQ
Skewness of Voltage
Median of Voltage
Charging Time CC
Skewness of Q
Diffusion coef
(c)
MAE
RMSE
PINN+ Virt. sensing
PINN
FNN
CNN
LSTM
2.20
2.84
3.07
3.95
3.62
4.39
4.90
5.92
5.23
6.16
Error [%]
(d-1)
PINN ($Q_{loss}$)
Upper Voltage [V]
Lower Voltage [V]
Trajectory MAE [%]
(d-2)
PINN with virtual sensing ($Q_{loss}$)
(e-1)
PINN (EOL)
EOL Error [EFCs]
(e-2)
PINN with virtual sensing (EOL)

Figure 7. Virtual sensing enables robust health assessment under V2G-restricted charging windows. (a) Capacity loss prediction under narrow high-voltage charging windows. (b) SHAP attribution for one multistep fast-charging profile within the narrowest voltage window [3.80V, 3.85V], corresponding to the typical ~50% SOC in practical V2G operation. (c) Comparison of purely ML baselines and the proposed PINN with and without virtual sensing under the [3.80V, 3.85V] charging window. (d, e) Performance of the proposed PINN across restricted charging windows using partial charging segments for (d) capacity loss and (e) EOL prediction. Bubble size and color denote the corresponding prediction error, where smaller and darker bubbles indicate improved performance.

**Discussion**

In this work, we propose a physics-informed learning framework with an integrated virtual-sensing strategy that infers key battery design parameters from short charging segments without additional sensors, including solid-state diffusion coefficient, negative electrode thickness, particle radius, and lithium-ion concentration. Digital-twin-derived aging knowledge is incorporated as soft constraints, with particle cracking representing a characteristic mechanism for Ni-rich NMC batteries under fast charging, while the remaining coupled processes are adaptively learned from data. As a result, accurate degradation trajectories and lifetime are recovered across a wide range of charging strategies and driving conditions using only 2% early-life observations. Compared with conventional ML approaches, the proposed framework provides two key advances. First, it opens the "black box" of physics-informed learning by visualizing constraint-guided loss landscapes, revealing how embedded mechanistic knowledge guides neural optimization. Second, virtual sensing extracts hidden design variables from existing charging signals, reducing trajectory prediction error by 40–50% and EOL prediction error by 5–7 fold. Under highly constrained charging windows, it further increases the fraction of windows with trajectory MAE below 3% from 81.0% to 85.7% and lifetime MAE below 100 EFC from 38.1% to 45.2%. More fundamentally, our findings show that effective physics-informed learning does not require fully resolved governing equations. Partially specified mechanisms, when jointly refined with data, provide sufficient constraints for reliable long-term extrapolation. Once trained, the framework replaces repeated multiphysics simulations with real-time computation, enabling rapid exploration of battery configurations tailored to usage requirements.

We believe this framework establishes a practical bridge between battery development and field deployment. By continuously translating charging behavior into interpretable design variables, it provides a transparent link between battery operation and health assessment, supporting advanced BMS strategies for emerging electrification scenarios, including V2G. Extending this framework to next-generation batteries and self-driving laboratories could enable application-driven discovery, transforming deployment from the endpoint of validation into a continuous driver of engineering development across energy-storage technologies and other complex multiphysics systems.

## Methods

### Particle cracking mechanisms fundamentals

The physical constraints, along with the first and second derivatives, are derived as follows:

$$Q_{loss} = A \times \frac{\psi}{Z} \times \left[ (1 + Z \times EFC_{norm} \times M)^{\frac{1}{B}} - (1 + Z)^{\frac{1}{B}} \right] \tag{1}$$

$$\frac{\partial Q_{loss}}{\partial EFC_{norm}} = \left( A \times \frac{\psi \times M}{B} \right) \times (1 + Z \times EFC_{norm} \times M)^{\frac{1}{B} - 1} \tag{2}$$

where

$$\psi = \frac{16\pi r_n{}^2 \rho_{cr} l_{cr,0} k \left( (\sigma_{t,max} b \sqrt{\pi a_0})^n {}_{SEI}^{inorganic}{}_{0}^{inorganic} \right)}{\delta_e M_{SEI}^{inorganic}} \tag{3}$$

$$Z = \frac{2-n}{2} k \left( (\sigma_{t,max} b \sqrt{\pi})^n a_0{}^{\frac{2}{2-n}} \right) \tag{4}$$

where A, B, $\Psi, Z$ are parameters affected by anode particle radius, diffusion-induced tangential stress, and material-related constants. The value of $M$ represents the maximum EFC in the training dataset and is used to normalize the EFC feature. Furthermore, in Equations. (3) and (4), the variable $r_n$ denotes the radius of the anode particle, and $\rho_{cr}$ represents the number of cracks per unit area on the graphite particle. The initial crack width is indicated by $l_{cr,0}$, and the initial crack depth by $a_0$. Over time, as aging progresses-represented by $N$-the initial crack depth $a$ increases according to material-specific constants $k$, $b$, and $n$. The maximum diffusion-induced tangential stress on the surface of the anode particle, caused by diffusion, is denoted by $\sigma_{t,max}$. The initial thickness of the inner inorganic SEI layer is $l_0^{inorganic}$, with its density and molecular weight given by $\rho_{SEI}^{inorganic}$ and $M_{SEI}^{inorganic}$, respectively. Additionally, the $\delta_e$ refers to the proportion of capacity loss attributed to the inner SEI layer relative to the total capacity loss for the exposed surface.

### Physics-informed learning with mechanistic constraints

Physics-informed neural network (PINN) integrates observational data with governing PDE constraints through automatic differentiation. Here, particle-cracking is embedded into the loss function as a constitutive aging constraint. The overall PINN loss combines (i) supervised measurement fitting, (ii) PDE residual minimization enforced through collocation points, and (iii) physics-consistency constraints, including initial condition anchoring and monotonic degradation regularization. Mathematically, the loss function $L$ can be represented as

$$L(\hat{Q}^{(\theta)}) = \lambda_d L_d + \lambda_{PDE} L_{PDE} + \lambda_{alloc} L_{alloc} + \lambda_{ic} L_{ic} + \lambda_{mono} L_{mono}, \tag{5}$$

where

$$L_d = \frac{1}{N_i} \sum_{i=1}^{N_i} \| NN(x_i, EFC) - Q_i \|, \tag{6}$$

$$L_{PDE} = \frac{1}{N_f} \sum_{j=1}^{N_f} \left\| \frac{\partial NN(x_j, t_{EFC,j})}{\partial t_{EFC,j}} - a \cdot \frac{\partial F_{phy}(x_j, t_{EFC,j}; \theta)}{\partial t_{EFC,j}} ) \right\|^2, \tag{7}$$

$$L_{colloc} = \frac{1}{N_f} \sum_{j=1}^{N_f} \| NN(x_j, t_{EFC}) - (a \cdot F_{phy}(x_j, t_j; \theta) + b) \|^2, \tag{8}$$

$$L_{ic} = \int_{\Gamma} \left\| \hat{Q}^{(\theta)}(x, 0) - Q_{ic}(x) \right\|^2 dx, \tag{9}$$

$$L_{mono} = \sum_{i=1}^{N} Re\,L\,U\left(\hat{Q}^{(\theta)}{}_i - \hat{Q}^{(\theta)}{}_{i+1}\right), \tag{10}$$

where $\hat{Q}^{(\theta)}$ denotes the predicted capacity loss function, $x$ represents features coordinates, $EFC$ represents cycle time coordinates. $\Omega$ represents the feature input domain, $\Gamma$ represents a time domain. $NN$ denotes the neural network operators. $Q_{ic}$ is initial conditions. and $\lambda_*$ denotes the weighting coefficients used to balance competing loss terms and stabilize PINN training.

In Equation (8), the loss function includes terms that ensure the neural network satisfies the initial conditions of the battery aging being modeled. The network's output at every collocation point must satisfy the physics (through the PDE residual) and conform to the known physical constraints (through initial conditions and monotonicity penalty). The monotonicity penalty $L_{mono}$ is introduced to enforce the physically expected degradation trend that capacity loss should not decrease with cycling, except for occasional capacity regeneration. Since there are still some unknown mechanisms triggering the aging behavior, especially when in extremely different conditions and wide operational range (0.6C-2C or MCC, 0degC-35degC, 70%DOD-100%DOD). Some cells show a linear or gradually decelerating decline in capacity fade, while other cells exhibit a transition from linear to accelerated degradation.

**Adaptive physics-informed transfer scheduler**

Traditional PINNs perform well for smooth and monotonic aging dynamics[45], but often struggle to capture highly coupled and non-smooth behavior using point-wise PDE residual minimization, which tends to produce over-regularized solution manifolds and steep optimization gradients. One limitation is that static physical constraints tend to bias the model toward smooth, long-lifetime trajectories, thereby missing abrupt aging transitions. An example can be seen in Figure S2 by the rapid degradation onset emerging near the 150th cycle at 0 °C when only sparse early-life data are available. These limitations motivate adaptive mechanistic refinement, where physical constraints evolve together with incoming operational signals rather than remaining globally fixed.

To reconcile incomplete constitutive physics with heterogeneous operating behavior, we introduce an adaptive physics-informed training scheduler tightly coupled with chronological transfer learning (see Supplementary Note 3 for details). The proposed scheduler continuously updates the constitutive parameters $\theta = \{A, B, \phi, Z\}$ together with the neural-network weights during target-cell adaptation. The optimization objective is formulated as:

$$W^*, \theta^* = argmin(L(\hat{Q}_{loss}{}^{(\theta)}))$$

where $L(\hat{Q}_{loss}{}^{(\theta)})$ combines measurement fitting, PDE residuals at collocation points, initial-condition anchoring and monotonicity regularization.

During adaptation, the scheduler operates across restricted V2G-relevant charging windows with lower bounds ranging from 3.50–3.80 V and upper bounds from 3.85–4.10 V (Section 2.4). Window-specific virtual-sensing descriptors extracted from partial charging segments are dynamically injected into the transfer process, enabling adaptive refinement under constrained sensing conditions. To preserve physically admissible parameter evolution, the constitutive variables are optimized through differentiable positivity reparameterization:

$$\theta_i = softplus(u_i) + \varepsilon$$

where $u_i$ denotes the unconstrained latent variable and $\varepsilon = 10^{-12}$ prevents numerical singularities. The scheduler shifts battery prognosis from fitting observed degradation to constraining physically plausible future evolution under incomplete observations.

**Physics-informed latent design space**

Virtual sensing is performed within a physics-informed latent design space comprising four design parameters ($L_{neg}$, $C_{s,max}$, $r_{neg}$, and $D_{neg}$). These parameters capture geometric and material characteristics governing battery electrochemical behavior and have been identified as the most informative representation for lifetime prediction, with additional design parameters providing diminishing predictive gains[4]. The latent design space is parameterized by a multiphysics DT calibrated through electrochemical characterization, including voltage/current profiles, EIS, IC and DV analysis, together with aging mode identification based on loss of lithium inventory and loss of active material[44]. Consequently, each point in the latent space represents a physically consistent battery design and generates a corresponding charging response through the DT. Virtual sensing projects partial charging signatures onto this latent design space using a random forest regressor trained on DT-generated synthetic data, enabling inference of four design parameters for downstream health assessment. The sensitivity ranking and uncertainty quantification of the selected design parameters are provided in Table S3 and Figure S31.

## Data and code availability

Data and code used in this work will be available after publication.

## Acknowledgements

The authors acknowledge "Best4Grid – Vehicle Battery Storage for Green Transport and Grid Stability in the Nordics", part of the Nordic Grand Solutions Programme funded by Nordic Energy Research, the Swedish Electromobility Center and the Swedish Energy Agency through the project "EV Drivers in the Driver's Seat." We also acknowledge support from the STandUP for Energy and Compel initiatives.